\documentclass[11pt,a4paper]{article}
\usepackage[nohyperref]{emnlp2018} %
\usepackage{times}
\usepackage{latexsym}
\usepackage{amsmath}
\usepackage{amssymb}
\usepackage{amsfonts}
\usepackage{linguex}
\usepackage{url}
\usepackage{tikz-qtree}
\usepackage{tikz-dependency}
\usepackage[textsize=tiny]{todonotes}
\usepackage{graphicx}
\usepackage{subcaption}
\usepackage{float}
\usepackage{array}
\usepackage{multirow}

\usepackage{verbatim} 

\aclfinalcopy 

\title{Under the Hood: Using Diagnostic Classifiers to Investigate and Improve how Language Models Track Agreement Information\\
~}

\author{
    Mario Giulianelli \\
    University of Amsterdam\\
    {\tt m.giulianelli@uva.nl\ \ } \\
    \\\And
    Jacqueline Harding \\
    University of Amsterdam\\
    {\tt hardingj@stanford.edu}\\
    \\\And
    Florian Mohnert\\
    University of Amsterdam\\
    {\tt mohnertf@gmail.com}
    \AND
    Dieuwke Hupkes\\
    ILLC, University of Amsterdam\\
   {\tt dieuwkehupkes@gmail.com} \\
    \\\And
    Willem Zuidema\\
    ILLC, University of Amsterdam\\
   {\tt w.h.zuidema@uva.nl} \\
   }

\begin{document}
\maketitle

\begin{abstract}
How do neural language models keep track of number agreement between subject and verb?
We show that `diagnostic classifiers', trained to predict number from the internal states of a language model, provide a detailed understanding of how, when, and where this information is represented. Moreover, they give us insight into when and where number information is corrupted in cases where the language model ends up making agreement errors. 
To demonstrate the causal role played by the representations we find, we then use agreement information to influence the course of the LSTM during the processing of difficult sentences. Results from such an intervention reveal a large increase in the language model's accuracy. Together, these results show that diagnostic classifiers give us an unrivalled detailed look into the representation of linguistic information in neural models, and demonstrate that this knowledge can be used to improve their performance.\\
\end{abstract}

\section{Introduction}

Machine learning models for estimating the probabilities of potential
next words (and hence, for predicting the next word) in a running text
have seen enormous improvements in performance over the last few years \citep{merity2018}. These newer models---all based on deep learning techniques such as LSTMs \citep{lstm1997}---allow some language technologies, such as speech recognisers, to reach `human parity'. 
From their high accuracy and from further analysis, it is clear that LSTM-based language models have learned a great deal about both short and long distance relations in sentences and discourse.
In particular, \citet{linzen2018} report that for several languages, their LSTM-based language model performs remarkably well on a set of long-distance number agreement tasks.

The Gulordava study, however, does not clarify which components of the LSTM are responsible for storing or processing syntactic features, and how such features are represented. Understanding how trained recurrent networks such as LSTMs might represent syntax and other structural information is currently a key area of research. Popular approaches include visualising the state space of these networks, performing ablations to the network, or using the internal states of the networks for some auxiliary task \cite[e.g.,][]{adi2016, kadar2017, conneau2018, khandelwal2018sharpfuzzy}.

In this paper, we analyse the phenomenon of subject-verb agreement in English using the \emph{diagnostic classification} approach of \citet{hupkes2018}.  
We start with replicating the results of \citet{linzen2018} on English, and we then show that diagnostic classifiers can be used to give a fine-grained analysis of how neural language models capture structural dependencies. 
In particular, we examine how information about subject-verb agreement is represented by an LSTM (Section~\ref{sec:dc}), (ii) how that information varies across timesteps (Section~\ref{sec:over_time}), and (iii) where and how the problems arise that let the model commit agreement errors (Section~\ref{sec:over_time}~and~\ref{sec:across_components}). 
Finally, to demonstrate how precisely and accurately this method can identify the network's internal representations, we (iv) show that we can alter the representation to strongly improve the model’s ability to predict verb number (Section~\ref{sec:improving}).
In the next section, after discussing subject-verb agreement, we outline the data used throughout our experiments.

\section{Data}\label{sec:data}

\begin{figure*}[t]
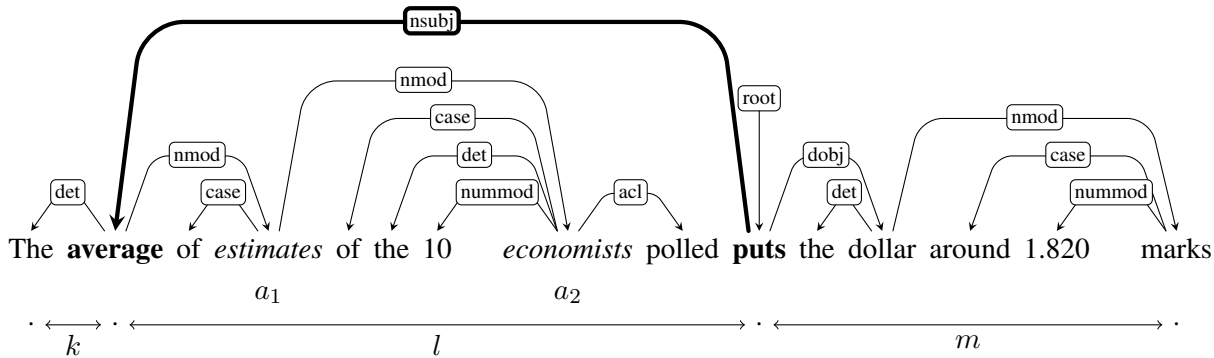

\centering
\begin{dependency} [edge slant=10pt]
    \begin{deptext}
    The \& \textbf{average} \& of \& \textit{estimates} \& of \& the \& 10 \&[.5cm] \textit{economists} \& polled \& \textbf{puts} \& the \& dollar \& around \& 1.820 \&[.5cm] marks\\
    \end{deptext}
    \deproot{10}{root}
    \depedge{2}{1}{det}
    \depedge[edge unit distance=2ex,{ultra thick}]{10}{2}{nsubj}
    \depedge{4}{3}{case}
    \depedge{2}{4}{nmod}
    \depedge{8}{5}{case}
    \depedge{8}{6}{det}
    \depedge{8}{7}{nummod}
    \depedge{4}{8}{nmod}
    \depedge{8}{9}{acl}
    \depedge{12}{11}{det}
    \depedge{10}{12}{dobj}
    \depedge{15}{13}{case}
    \depedge{15}{14}{nummod}
    \depedge{12}{15}{nmod}
    \node (subject) [below of = \wordref{1}{2}, xshift = 0cm] {$\cdot$};
    \node (verb) [below of = \wordref{1}{10}, xshift = 0cm] {$\cdot$};
\node (bottomleft) [below of = \wordref{1}{1}, xshift = 0cm] {$\cdot$};
\node (bottomright) [below of = \wordref{1}{15}, xshift = 0cm] {$\cdot$};
\node (attractor1) [below of = \wordref{1}{4}, yshift = .4cm] {$a_1$};
\node (attractor2) [below of = \wordref{1}{8}, yshift = .4cm] {$a_2$};
\draw [<->] (bottomleft) -- (subject) node[midway,below] {$k$};
\draw [<->] (subject) -- (verb) node[midway,below] {$l$};
\draw [<->] (verb) -- (bottomright) node[midway,below] {$m$};
\end{dependency}
\caption{An example dependency parse of a sentence with a singular subject head and main verb (marked in boldface). As the subject \textit{average} and the verb \textit{put} are separated by 7 tokens, the \textit{context size} ($l$) of this sentence is 7. Within this context, there are two intervening plural nouns, \textit{estimates} ($a_1$) and \textit{economists} ($a_2$), which we call \textit{agreement attractors}.}
\label{fig:dep-tree}
\end{figure*}

The work in this paper focuses on understanding how recurrent neural language models can understand subject-verb agreement, which is used as a proxy for understanding syntactic structure.
In this section, we discuss subject verb agreement and the type of sentences we look at throughout the rest of this paper.
We then briefly describe the data that we use for our investigation. 

\subsection{Subject-verb agreement}
Subject-verb agreement is a variable-distance syntactic dependency, and a classic example of a structural dependency in natural language \citep{chomsky1957,tesniere1959}.
In English, a present tense verb and the head of its syntactic subject must agree on their number (singular or plural). Thus, ``The \textbf{dog} \textbf{chases} the cat'' is grammatical, whilst ``The \textbf{dog} \textbf{chase} the cat'' is not. 
In principle, subject and verb can be separated by an arbitrary number of tokens, often including other nouns with a potentially different number (for an example, see Figure~\ref{fig:dep-tree}).
We call the number of tokens between the subject head and the mean verb the \emph{context size}.

Without any syntactic analysis, it is unclear how to identify all subject-verb pairs in a sentence within an arbitrarily large window of tokens, especially since intervening nouns can themselves be candidates for agreement.
To respect subject-verb agreement, a language model needs to detect the grammatical number of both the subject head and the verb, store this information across timesteps, and identify which nouns correspond to which verbs.
When intervening nouns carry the opposite grammatical number from the subject head---as do both intervening nouns in the example sentence in Figure~\ref{fig:dep-tree}---we refer to them as \textit{agreement attractors}, or simply \textit{attractors}.   

\subsection{Datasets}\label{sec:extraction}
For the experiments described in this paper we use two different datasets. 
The first is the one introduced by \citet{linzen2018}, which contains 410 sentences with at least three tokens occurring between subject head and verb. 
For each of 41 original sentences, nine `nonce' variants were generated by substituting each context word in the sentence by a random word with the same part-of-speech tag and morphological features. 
This data construction method is motivated by the fact that grammaticality judgements should not be influenced by the meaningfulness of a sentence, and ensures that frequency-based confounds are avoided.
Every sentence in the dataset is annotated with the correct and incorrect verb forms, the morphological features of the former, the position of the subject head and of the verb, the number of agreement attractors, and the type of construction spanning the long-distance dependency. 

Additionally, we extract different subsets from a corpus with number prediction problems extracted from an annotated Wikipedia corpus by \citet{linzen2016}.\footnote{\url{https://github.com/TalLinzen/rnn_agreement}}.
The large amount of annotated sentences in this dataset (ca.\ 1.5 million) allows us to retrieve sets of sentences that satisfy specific conditions relevant to subject-verb agreement.
In particular, we can extract sentences with specific context sizes, and fixed numbers of words before the subject and after the verb.
We are also able to specify whether the sentences in the set should have an attractor and---if so---at which index (or, in our terminology, \emph{timestep}) the attractor should appear.
Similarly, we can ensure that there is no other noun between subject and verb that has the same number as the subject (we call these \emph{helpful nouns}). 
As we will see, this allows us to examine the dynamic effect of attractors in the way the LSTM processes subject-verb agreement. 

In this paper, the specific subset of the universal dependency dataset we use varies from experiment to experiment, as different experiments require different constraints. 
We will specify our selection of data for each experiment in the relevant sections. 
To clarify which subset of the wikipedia dependency corpus is used in an experiment, we use the following notation: \emph{WD-Kk-Ll-Mm-Aa}, where \emph{k} refers to the minimal number of words appearing before the subject, \emph{l} to the number of words between the subject and verb (the context size), \emph{m} to the minimal number of words after the verb, and \emph{a} to the position of the attractor relative to the subject.
We use an asterisk to indicate that no restrictions are placed on one of the above mentioned variables; e.g., \emph{A*} indicates that there may or may not be an attractor.
Finally, we denote datasets of sentences that have no attractor with a minus following the attractor index (i.e., \emph{A$-$}).

\section{Replication}\label{sec:replication}

We start with replicating the experiment performed in \citep{linzen2018}, using the pre-trained LM and the English test set made available by the authors of the paper.\footnote{\url{github.com/facebookresearch/colorlessgreenRNNs/tree/master/data}}
Following \citet{linzen2016} and \citet{linzen2018}, we use the LSTM language model to process a corpus of sentences containing long-distance subject-verb relations, and test whether the model assigns a higher probability to the verb that originally occurred in the sentence than to its incongruent counterpart.

\begin{table}[ht!]
    \centering
    \begin{tabular}{c | c | c }
        &\textbf{\citeauthor{linzen2018}} & \textbf{Our Accuracy} \\
        \hline
        Original   &  81.0  &  78.1 \\
        Nonce      &  74.1  &  70.7 \\
    \end{tabular}
    \caption{LM accuracy on both English sets from \citet{linzen2018}. Reported are the percentages of sentences for which the correct verb form is assigned a higher likelihood under the LM than the incorrect form.
    }
    \label{tab:Linzen_replication}
\end{table}

In Table~\ref{tab:Linzen_replication} we report both Gulordava's original accuracies, and the results from our replication. 
Overall we obtain similar results, but our accuracy scores are slightly lower\footnote{The results we obtain with our implementation exactly match those we get when running the script publicly shared by \citet{linzen2018}; we currently have no explanation for the discrepancy in overall scores but consider the differences small enough to proceed with the real purpose of our study: understanding how the models work.}  than those reported by \citet{linzen2018}.

\section{Diagnostic Classification to Predict Number}\label{sec:dc}

\begin{table*}[hbt]
\centering
    \begin{tabular}{c|ccccc}
    & $\mathbf{h_t}$ & $\mathbf{c_t}$   & $\mathbf{f_t}$  & $\mathbf{i_t}$   & $\mathbf{o_t}$ \\
    \hline
   \textbf{Layer 0} & \textbf{0.74} / 0.57 & \textbf{0.76} / 0.58 & \textbf{0.69} / 0.55 & \textbf{0.68} / 0.56 & \textbf{0.69} / 0.56 \\
   \textbf{Layer 1} & \textbf{0.90} / 0.62 & \textbf{0.91} / 0.65 & \textbf{0.86} / 0.61 & \textbf{0.86} / 0.60 & \textbf{0.87} / 0.60 \\
    \end{tabular}
    \caption{Mean accuracy of DCs (correct/wrong) across timesteps, averaged over datasets drawn from different context sizes and attractor positions (with $K\!=\!0$, $M\!=\!0$, $5\!\leq\!L\!\leq\!7$ and with a variable number of attractors at different positions).}
    \label{tab:dc_overall_accuracies}
\end{table*}

\begin{figure*}[htb]
\centering
\includegraphics[width=\linewidth, trim=0mm 0mm 95mm 0mm, clip]{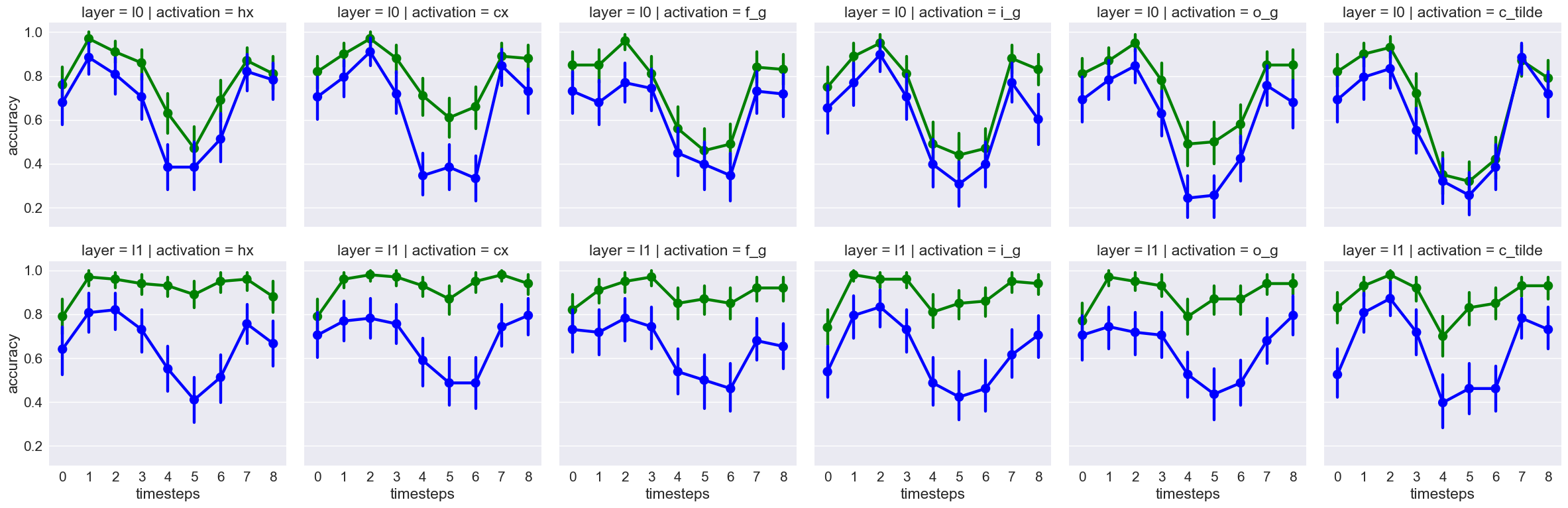}
\caption{Accuracies over time (on WD-K1-L5-M1-A3) of 10 diagnostic classifiers trained and tested on data from different components of the LSTM.
    As in this testset one word occurs before the subject, the subject is at timestep 1.
Green lines represent sentences for which the LSTM predicts the correct verb, blue lines sentences for which the LSTM assigns a higher probability to the incongruent counterpart.}
\label{fig:cs_5_attractor_idx_3_classification}
\end{figure*}

After confirming \citet{linzen2018}'s results, we now investigate \emph{how} the LSTM represents the required number information, how this information is built up over time and where in the network the representation resides.
To this end, we use diagnostic classifiers \cite[DCs,][]{hupkes2018}.
The key idea of diagnostic classification is to test whether an LSTM's intermediate representations contain information about a particular phenomenon---such as subject-verb agreement---by training another model to recognise the information relevant to the phenomenon in the internal activations of the LSTM. 
More precisely, given a dataset of intermediate LSTM representations and a set of labels that describe the hypothesis to be tested, a meta model can be trained to predict the correct label from the representations.
If the model succeeds in this task (i.e.\ if it achieves a performance significantly above chance on test data disjoint from the training data), this constitutes evidence that the LSTM is in fact computing or keeping track of the hypothesised information.

\paragraph{Training}
We create a training set containing 1000 sentences that all have 5 words between subject and verb (i.e.\ the context size is 5), have at least one word before the subject and after the verb, and for which no attractor based constraints are placed on the training set (\textit{WD-K1-L5-M1-A$-$}).
We run the pretrained LM of \citet{linzen2018}---a two layer LSTM model with 650 hidden units---on this corpus, and for both layers we extract activation data for both the hidden and gate activations (the hidden activation $\mathbf{h_t}$ and memory cell $\mathbf{c_t}$, and the forget gate $\mathbf{f_t}$, input gate $\mathbf{i_t}$ and output gate $\mathbf{o_t}$).
For example, for a single sentence of length $n$
we obtain $5 \times 2 \times n$ activation vectors, because we have 2 layers, \textit{n} timesteps, and 5 types of activations at each time step $t$: $\mathbf{h_t}, \mathbf{c_t}, \mathbf{f_t},\mathbf{i_t},\mathbf{o_t})$. 
We then label all activations with the number of the main verb of the sentence from which it was generated (either `singular' or `plural') and train a separate DCs for each of the 10 components of the LSTM.

\paragraph{Results}
We test the trained DCs on two test sets, that differ with respect to whether the LM correctly or incorrectly classified the sentences they contain (i.e.\ a sentence $s$ is in the `correct' set iff the LM assigns higher probability to the correct form of the sentence than to the incorrect form). 
Otherwise, the two sets have similar features, containing both sentences from \emph{WD-K1-L5-M1-A3}. 
While we strive to generate the `wrong' and `correct' test sets with 100 sentences each, this is not always possible due to data sparsity. 
However, we ensure that both test sets have approximately the same size and do contain at least 50 sentences.

In Table \ref{tab:dc_overall_accuracies}, we print the average DC accuracies.
We observe that for both the `wrong' and the `correct' test sets, the accuracies are highest at the second layer (layer 1) across almost all LSTM components, suggesting that the last LSTM layer reaches the level of abstraction which can best capture long-distance dependencies. 

In Figure \ref{fig:cs_5_attractor_idx_3_classification}, we plot the average DC accuracy at different timesteps when processing sentences (from a set with a context size of 5 and a single attractor located three words after the subject). 
Unsurprisingly, the DCs obtain their best accuracy scores at (or just after) the subject and verb timestep.
This pattern is consistent across context sizes, attractor positions, and number of words before the subject and after the verb, and regardless of whether the LSTM prediction was correct or incorrect.
This result illustrates that the LM learns to recognise the number information of subject heads and present tense verbs. 

The figure furthermore shows that performance differs between layers and between components. The DC performance of the layer 1 components, moreover, critically differs for `correct' and `wrong' sentences, 
For example, classifiers that make predictions based on $\mathbf{c_t}$ and $\mathbf{h_t}$ activations of `correct' sentences are the most stable in terms of accuracy, in particular at layer 1. 
Although all LSTM components outperform the random baseline of 50\%, these results imply that the cell state and the hidden activation are the LSTM components that are most specialised at processing number information. 
We test this claim in Section \ref{sec:over_time}. 

Another cause of differences across diagnostic classification error rates is the presence of agreement attractors. 
Accuracies for the test sets with an attractor are overall lower than those obtained on sentences without an attractor. 
While the error rate rises in Figure \ref{fig:cs_5_attractor_idx_3_classification} and diverges between `correct' and `wrong' at the position of the attractor, the same does not happen for sentences without attractors (not plotted).

\section{Representations Across Timesteps}\label{sec:over_time}

Results so-far show us that number information is most easily retrieved from the internal states of the LM when the noun or verb have just been presented, but not very well from the internal states at intermediate timesteps. The good performance of the LM in predicting the  number of the verb, however, indicates that the LM does retain the subject's number information during those intermediate timesteps---but apparently it does so using a \emph{different} representation. 
In this section, we focus on these changing representations.

\begin{figure}[h!]
    \includegraphics[width=0.86\linewidth, trim= 10mm 0mm 40mm 10mm, clip]{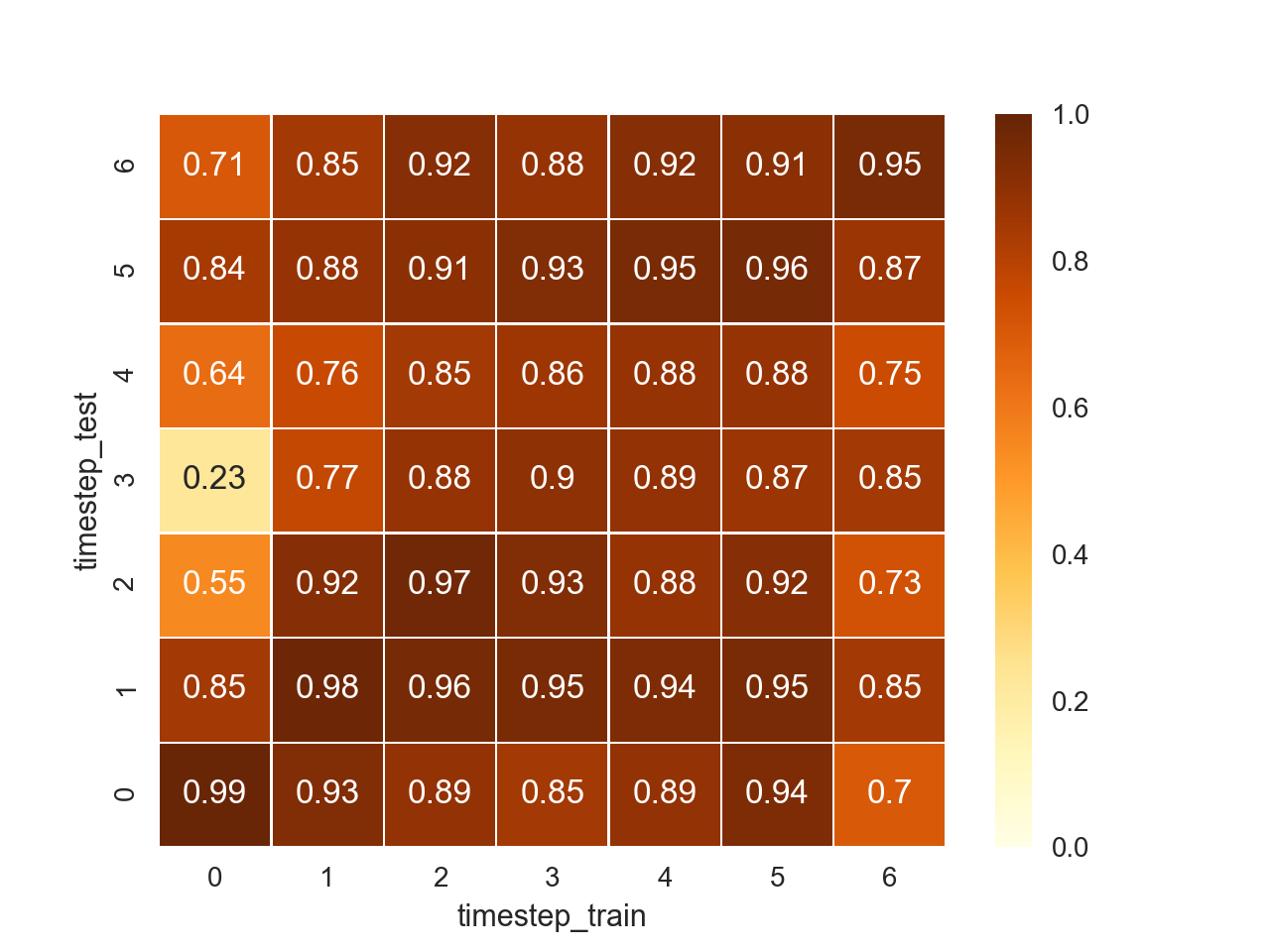}
    \includegraphics[width=\linewidth, trim= 10mm 0mm 20mm 10mm, clip]{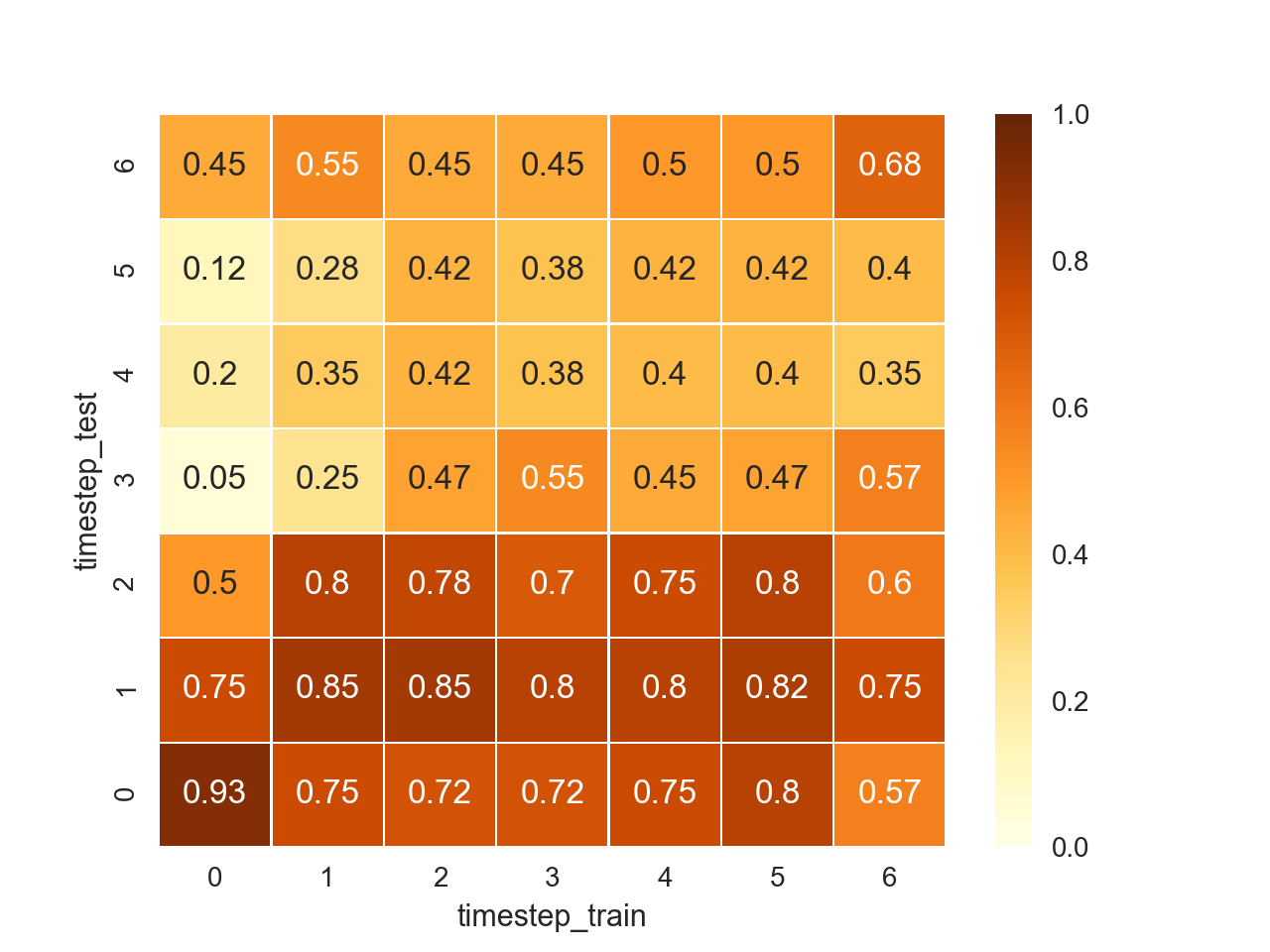}
    \caption{The temporal generalization matrices for DCs trained on memory cell activation at different timesteps, for correctly (top) and incorrectly classified (bottom) sentences. Timestep 0 corresponds to the subject of the sentence, the attractor and main verb of the sentence occur at timesteps 3 and 6, respectively. The corpus used for testing here is \emph{WD-K*-L5-M*-A3}.}
    \label{fig:timestep_heatmap}
\end{figure}

In the previous experiment we trained diagnostic classifiers on activation data for all words in the sentence. 
In contrast, we now train \emph{separate} diagnostic classifiers for each timestep: each DC$_t$ is trained with activation data at timestep $t$ only. 
We test, however, each DC$_t$ on data from all other timesteps as well. 
With a total of $T$ timesteps, this gives us $T \times T$ DC-accuracies that together constitute a \emph{Temporal Generalization Matrix} \citep{king2014matrix,fyshe2016brain}. 

In effect, we are forcing each DC to specialise on timestep-specific representations of subject-verb agreement information. 
If this information is represented uniformly across timesteps, a classifier trained at the subject timestep should also have a high accuracy when applied to the activations corresponding with the timestep in which the attractor occurs.
If, on the other hand, information is dynamically encoded, no such generality of classifiers is to be expected.

\paragraph{Data}
To test the development of the encoding over time, we create a corpus with sentences that are identical with respect to the position of the subject, attractor and main verb. 
We train on sentences with 5 intervening words between the subject, containing one attractor 3 timesteps after the subject, and a variable number of words before the subject and after the verb (\emph{WD-K*-L5-M*-A3}).
After computing the activations for all sentences, we collect the activations corresponding to all 6 timesteps from subject to verb, in 6 different bins. For each bin, we train a separate DC.

For testing we create again a `correct' and an `incorrect' test set, drawing both sets from \emph{WD-K*-L5-M*-A3}. 
Following the same procedure as for the training data, we split both test sets up into 6 timesteps.
In the remainder of this section, position 0 thus always refers to the position of the subject, while the attractor and main verb of the sentence occur at timestep 3 and 6, respectively.

In Figure \ref{fig:timestep_heatmap}, we plot the Temporal Generalization Matrix for the memory cell ($\mathbf{c}^1_t$) activation data, containing the accuracies of $T$ DC's evaluated on $T$ timestep datasets each. The top figure plots results for `correct' sentences, the bottom figure for `incorrect' sentences. 

A first observation is that accuracies on the diagonals---which correspond to classifiers that were trained and tested on the same timestep---are typically high for sentences that are processed correctly, while being lower for incorrectly processed sentences.
Interestingly, this difference already emerges at the first two timesteps, where no attractor has yet appeared---suggesting that an important part of the problem with misclassified sentences is the encoding of the relevant information already when the subject occurs.

Comparing the plots for correctly and incorrectly processed sentences, we notice that the attractor (timestep 3) has a very large effect on the accuracies for incorrectly classified sentences. For those sentences, the LM's internal states contain no information anymore after the attractor is processed: timesteps 4 and 5 receive below chance accuracies, whereas for correctly processed sentences the attractor prompts only a slight dip in accuracy. 

Focussing on the correctly processed sentences, an interesting observation that can be made is the discrepancy between column 0 and 6 (the columns corresponding to the subject and verb of a sentence) and the rest of the columns. While the first and last column generalise poorly to different timesteps, the classifiers trained and tested on timesteps 1-5 show a different pattern: despite potential effects from the attractor at timestep 3, the accuracy scores do not change substantially across timesteps.
This implies that the LSTM represents subject-verb agreement information in at least two different ways: a short-term `surface' level at and around the subject timestep, and a longer-term `deep' level for successive sequence processing.
This deep level information seems to be represented most generically at timestep 4, the classifier for which has the highest accuracy across timesteps.

In the next section, we delve deeper into the representations at this timestep and investigate which components of the LSTM are most crucial in representing this information.

\section{Comparing Representations Across Components}\label{sec:across_components}

In this section, we briefly investigate the stability of information across components of the LSTM.
Rather than comparing DCs that are trained on different \emph{timesteps}, we now compare DCs that are trained on different \emph{components}.
We focus on timestep 4 which, following our previous experiments, optimally represents `deep' information about subject-verb agreement.
For our experiments, we use the same training set as for the previous experiment, with sentences with a context size of 5 and a single attractor located three words after the subject (\emph{WD-K*-L5-M*-A3}).

\begin{figure}[t]
    \includegraphics[width=1.05\linewidth, trim= 6mm 20mm 5mm 2mm, clip]{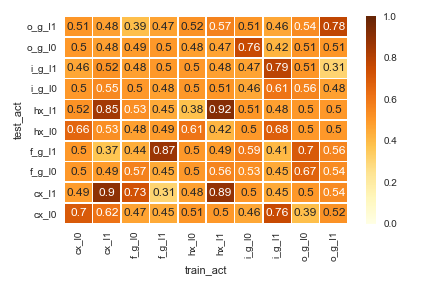}
    \includegraphics[width=1.05\linewidth, trim= 6mm 2mm 5mm 2mm, clip]{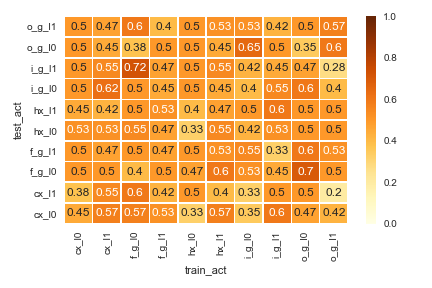}
    \caption{The spatial generalization matrices at timestep 4. Shown are accuracies of DCs trained on activation data of each component separately (horizontal), and tested on each component separately (vertical). Results for correctly (top) and incorrectly (bottom) classified sentences.}
    \label{spatialgeneralizationmatrix}
\end{figure}

Figure~\ref{spatialgeneralizationmatrix} presents the `spatial generalization matrix', with DCs trained at timestep 4 with data from each components separately. The matrix shows that deep information is best represented in the hidden activation and memory cell of layer 1, and that the representations in these two components are similar.

\begin{figure*}[t!]
\includegraphics[width=\linewidth, trim = 0mm 0mm 0mm 0mm, clip]{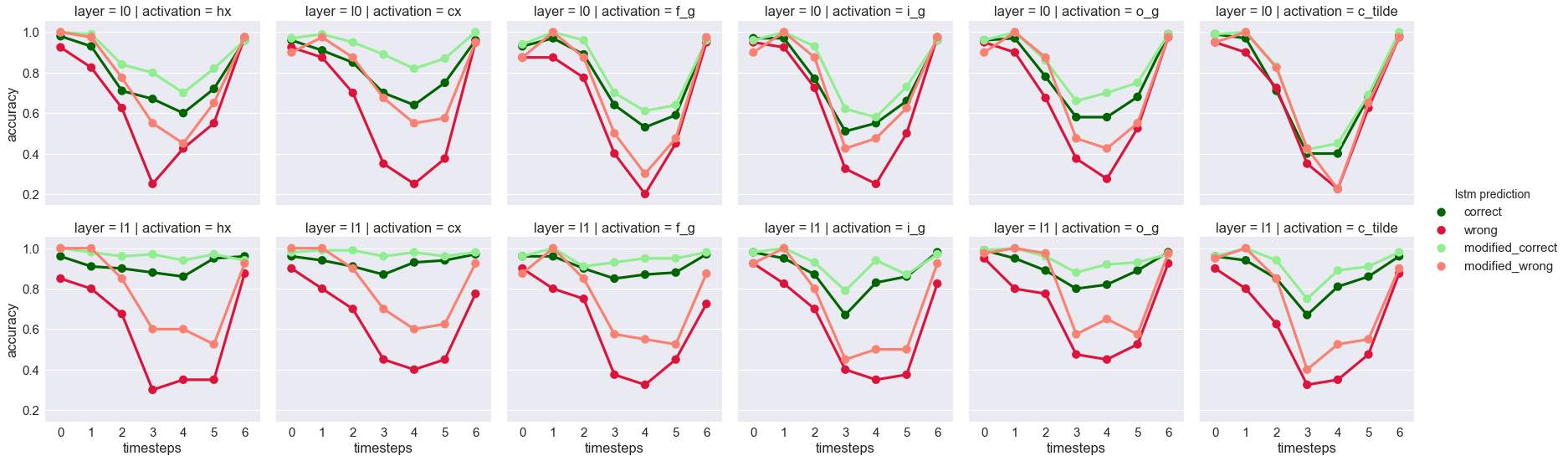}
\caption{Mean accuracies for each component of the LSTM after an intervention of $\mathbf{c_t}$ and $\mathbf{h_t}$ at the subject timestep 0. An attractor and the agreeing verb occur at timestep 3 and 6, respectively.}
\label{fig:intervention-subj}
\end{figure*}

\begin{table*}[bt]
\centering
\begin{tabular}{l|llllll|l|l} \hline
&
An&
official& 
estimate& 
issued&   
in     &  
2003   &  
suggests&   
suggest\\
Original &&
-11.05&
-8.426&
-8.472&
-1.243&
-3.951&
-5.753&
-5.6979\\
Intervention &&
-11.05&
-8.426&
-8.472&
-1.268&
-3.97  &   
-5.691&
-6.4361\\ \hline
\end{tabular}
\caption{Example sentence as processed by the neural language model of \citet{linzen2018}, without and with our intervention. Shown are perplexities per word, for two versions of the sentence (featuring the verb `suggests' or `suggest').}
\label{examplesentence}
\end{table*}

\section{Improving the Language Model Using Diagnostic Classifiers}\label{sec:improving}

In the experiments presented above, we used diagnostic classifiers to investigate the way the LSTM performs the verb number prediction task. 
In this section, we take one step further: rather than using DCs to analyse what neural networks are encoding, we try to use them to actively influence their behaviour through what they learned.

We use the same data as we used for the experiments described in the previous section: a corpus of sentences with the subject at timestep 0, one attractor 3 timesteps after (at timestep 3) and the main verb at timestep 6 (\emph{WD-K0-L5-M0-A3}).
We train 4 DCs to predict the number of the sentence from the hidden layer activations and memory cell activations for both layers, respectively.

We then use the trained DCs to actively influence the course of processing by the LSTM. We start processing sentences from the \citet{linzen2018} corpus, but after processing the subject of a sentence---the point where we discovered information is stored in a corrupted way for `wrong' sentences---we halt the LSTM's processing, extract the hidden activation and the activation of the memory cell, and apply the trained diagnostic classifier to predict whether the main verb in the sentence is singular or plural.
We then slightly adapt the activations based on the error that is defined by the difference between the predicted label and the correct label for this particular sentence. We compute the gradients of this error with respect to the activations of the network, and we modify the activations using the delta-rule (we empirically decided on $\eta=0.5$).
In other words, we change the activations such that the prediction of the diagnostic classifier is slightly closer to the gold label.
After adapting the activations, we continue to process the rest of the sentence given the adapted activations.

\paragraph{DC accuracy} In Figure \ref{fig:intervention-subj} we plot the accuracies of DCs trained on different components of the LSTM when we apply them on activations resulting from sentences processed with the above described intervention.
Trivially, the intervention increases the accuracy of DCs for the hidden activation and memory cell of the network at timestep 1.
More interestingly, this effect persists while the processing of the sentence proceeds---in some cases it grows even stronger---and thus in fact \emph{changes} how the LSTM processes the sentence.
This effect is not only visible in the components on which the intervention is done, but also displays in the gate-values, that are not directly updated but only changed indirectly through the interventions in the memory cell and hidden activations.

\begin{table}[h!]
    \centering
    \begin{tabular}{c | c | c }
        & \textbf{without} & \textbf{with} \\ 
        & \textbf{intervention} & \textbf{intervention}\\
        \hline
        Original   &  78.1  &  85.4 \\
        Nonce      &  70.7  &  75.6 \\
    \end{tabular}
    \caption{Accuracy of the LSTM on the \citet{linzen2018} agreement test, with and without an intervention at the subject timestep.} 
    \label{tab:replication_with_intervention}
\end{table}

\paragraph{Language modelling}
To put our interventions to the test, we now assess the predictions made by the LSTM as a consequence of the interventions.
First, we confirm that the intervention does not cause strong anomalies in the LSTM, by comparing the perplexity of a small corpus of sentences processed \textit{with} interventions at the subject timestep with sentences processed without any interventions. Table~\ref{examplesentence} shows an example sentence.
We do not find any strong differences, confirming that the intervention is minor with respect to the overall behaviour of the LSTM. 
On the agreement test described by \citet{linzen2018} and conducted earlier in Section \ref{sec:replication}, however, the intervention \textit{does} have a strong effect, as can be seen in Table \ref{tab:replication_with_intervention}.
The accuracy of predicting the correct verb number increases from 78.1 to 85.4 and from 70.7 to 75.6 for original and nonce sentences, respectively.

These results provide evidence that DCs are able to pick up features that are actually used by the LSTM, rather than relying on idiosyncrasies in the high dimensional spaces that happen to be aligned with the predicted labels.
Furthermore, they illustrate how diagnostic classifiers can be used to actively change the course of processing in a recurrent neural network, and with this opens a path that moves from merely \textit{analysing} to actively \textit{influencing} black box neural models.

\section{Conclusions}

In this paper, we focus on understanding how an LSTM language model processes subject-verb congruence, using a task first presented by \citet{linzen2016}, in which it is tested whether a language model prefers congruent over incongruent verbs.
After replicating their results, we train diagnostic classifiers \cite{hupkes2018} to discover where and how number information is encoded by the LSTM; we find that number information is encoded \emph{dynamically} over time, rather than remaining constant.
Using a cognitive-neuroscience inspired method, we then train different diagnostic classifiers for different timesteps, resulting in a \emph{Temporal Generalisation Matrix}, which provides more information about changing representations over time.
We find that while number information is stored in very different ways at the beginning and end of a sentence, in between a relatively stable `deep' representation is maintained.
Additionally, we find that for sentences in which the LSTM prefers an incongruent verb over congruent one, the information appears to be stored wrongly already at the beginning of the sentence, far before the verb is to appear.

Combining this information, we invert the process of diagnostic classification, using the classifiers to \emph{influence} rather than merely observe.
To this end, we process sentences with our language model and, at the point where we find information to be often corrupted, we intervene by (slightly) changing the hidden activations of the network using a trained DC. After this intervention, we continue processing the sentence as normal.
This small intervention has little effect on the overall course of the LSTM, but a very large effect on the verb prediction at the end: the percentage of sentences for which the model prefers the congruent over the incongruent verb rises from 78.1\% to 85.4\%.

With these results, we not only show that diagnostic classifiers offer a detailed understanding of where and when information is encoded in a neural model, but also that this information can be used post hoc to change the course of the processing of such a model.

\section*{Acknowledgements}

DH and WZ are funded by the Netherlands Organization for Scientific Research (NWO), through a Gravitation Grant 024.001.006 to the Language in Interaction Consortium.

\bibliography{emnlp2018}
\bibliographystyle{acl_natbib_nourl}

\end{document}